\def\conference{0}
 \def\shownotes{1}
\def\membershipinference{1}

\ifnum\conference=1

\documentclass[10pt,twocolumn,letterpaper]{article}

\AtBeginDocument{%
  \providecommand\BibTeX{{%
    \normalfont B\kern-0.5em{\scshape i\kern-0.25em b}\kern-0.8em\TeX}}}

\usepackage{balance}
\else
\documentclass[10pt]{article}
  \usepackage[top=3cm, bottom=3cm, left=2cm, right=2cm]{geometry} 
\fi

\usepackage{times}
\usepackage{epsfig}
\usepackage{graphicx}
\usepackage{amsmath}
\usepackage{amssymb}
\usepackage[normalem]{ulem}
\usepackage{setspace}
\usepackage{times}
\usepackage{epsfig}
\usepackage{amsmath}
\usepackage{amssymb}
\usepackage{makecell}
\usepackage{verbatim}
\usepackage{balance}
\usepackage{enumitem}
\usepackage{color}
\usepackage{csquotes}
\usepackage{cite}
\usepackage{xspace}
\usepackage{enumerate}
\usepackage{graphicx}
\usepackage[noend]{algpseudocode}
\usepackage{algorithm}
\usepackage{wrapfig}
\usepackage{microtype}
\usepackage{pifont}
\usepackage{caption}
\usepackage{multirow}
\usepackage{hyperref}
\DeclareMathOperator*{\argmin}{arg\,min}

\usepackage{csquotes}

\ifnum\shownotes=1
\newcommand{\authnote}[2]{{\textcolor{red}{\textsf{#1 notes: }\textcolor{blue}{#2}}\marginpar{\textcolor{red}{\textbf{!!!!!}}}}}
\else
\newcommand{\authnote}[2]{}
\fi

\newcommand{\upslope}{\ensuremath{\mathtt{Upslope}}\xspace}
\newcommand{\update}{\ensuremath{\mathtt{Update}}\xspace}
\newcommand{\downslope}{\ensuremath{\mathtt{Downslope}}\xspace}
\newcommand{\rand}{\ensuremath{\mathtt{Rand}}\xspace}
\newcommand{\srand}{\ensuremath{\mathtt{SR}}\xspace}
\newcommand{\concat}{\ensuremath{\mathtt{Concat}}\xspace}
\newcommand{\sral}{\ensuremath{\mathtt{SRwAL}}\xspace}

\begin{document}

\ifnum\conference=0
\title{Privacy Attacks Against  Biometric Models with Fewer Samples: Incorporating the Output of Multiple Models}

\else
\title{Inverting Biometric Models with Fewer Samples: Incorporating the Output of Multiple Models}
\fi

\author{Sohaib Ahmad\\
University of Connecticut\\
Storrs\\
{\tt\small sohaib.ahmad@uconn.edu}
\and
Benjamin Fuller\\
University of Connecticut\\
Storrs\\
{\tt\small benjamin.fuller@uconn.edu}
\and
Kaleel Mahmood\\
University of Connecticut\\
Storrs\\
{\tt\small kaleel.mahmood@uconn.edu}
}

\maketitle
\begin{abstract}

Authentication systems are vulnerable to model inversion attacks where an adversary is able to approximate the inverse of a target machine learning model. Biometric models are a prime candidate for this type of attack. This is because inverting a biometric model allows the attacker to produce a realistic biometric input to spoof biometric authentication systems.

One of the main constraints in conducting a successful model inversion attack is the amount of training data required. In this work, we focus on iris and facial biometric systems and propose a new technique that drastically reduces the amount of training data necessary. By leveraging the output of multiple models, we are able to conduct model inversion attacks with $1/10$th the training set size of Ahmad and Fuller (IJCB 2020) for iris data and $1/1000$th  the training set size of Mai et al. (Pattern Analysis and Machine Intelligence 2019) for facial data. We denote our new attack technique as \textit{structured random with alignment loss}.





Our attacks are black-box, requiring no knowledge on the weights of the target neural network, only the dimension and values of the output vector. 

 \ifnum\membershipinference=1
 To show the versatility of the alignment loss, we apply our attack framework to the task of membership inference (Shokri et al., IEEE S\&P 2017) on biometric data. For the iris, membership inference attack against classification networks improves from $52\%$ to $62\%$ accuracy. 
 \fi
 

\end{abstract}






\graphicspath{images/}
\graphicspath{ {images/} }

\section{Introduction}
\ifnum\conference=0
Biometric identification is widely used, most devices contain multiple biometric modalities.  Most prior research focuses on the accuracy of identification models, the privacy implications of these models are not clearly understood.   Apple's documentation\footnote{https://support.apple.com/en-us/HT208108.} on FaceID states devices store ``mathematical representations of your face.''  The goal of this work is to understand the privacy risks of leaking the model output (the mathematical representations).

\fi
Many authentication systems are based on biometric identification~\cite{ratha2001enhancing,jain2007handbook}.  Two widely adopted biometrics include iris and facial recognition. Despite the prevalence of these biometric based authentication systems, they remain vulnerable to a type of attack called a model inversion~\cite{fredrikson2015model}. In a model inversion attack, an adversary is able to train an attack model that approximates the inverse of the target biometric model used in the authentication system. Once the adversary is able to succeed in training this attack model, they are able to produce realistic looking biometrics. These realistic looking biometrics can be used for spoofing attacks~\cite{marcel2014handbook}, where an attacker creates a ``fake'' version of a user's biometric. 

Deep learning models are increasingly being used for biometrics~\cite{deng2019arcface,liu2017sphereface,wang2018cosface,wang2019cross,zhao2017towards,ahmad2019thirdeye}.  Fredikson et al.\,initiated model inversion attacks on such networks, targeting the facial biometric~\cite{fredrikson2015model}. Recent model inversion attacks use generative adversarial networks or GANs~\cite{zhang2020secret} and use auxiliary information such as blurred faces.


 To set notation, denote the trained biometric identification system as $f_T$ to indicate it is the model being targeted in the attack.  The attack proceeds in stages:
 \begin{description}
 \item[Training] The attacker receives $\ell$ samples of the form \[(x_i, y_i = f_{T}(x_i)).\]  At the end of this stage the attacker outputs a model $f^{-1}_T$.  It should be the case that for unseen pairs $x', y'$ the value $f^{-1}_T(y')$ is similar to $x'$. 
 \item[Test/Attack] The attacker receives values $y'$ and inverts them to produce realistic biometric values $f^{-1}_T(y')$.
 \end{description}

\noindent 
A limitation of prior work is the need for a large number of training samples. 
Mai et al.~\cite{mai2018reconstruction} require $2\times 10^6$ training samples in their attack on the facial biometric. Ahmad and Fuller~\cite{ahmad2020resist} require $2\times 10^4$ training samples in their attack on the iris biometric.  While large facial and iris datasets exist, model inversion targets smaller applications.  It is thus crucial to determine if model inversion is possible with fewer training points. 

We investigate whether the adversary can substitute the output of multiple models in \textbf{Training} in place of more training samples.
\ifnum\conference=1\footnote{Salem et al.~\cite{salem2020updates} study the difference a model undergoes when it is updated in an online fashion. Their work considers small updates while we explore larger changes when the target model's dataset undergoes deletion or addition of classes.} 
\else
Salem et al.~\cite{salem2020updates} study the difference a model undergoes when it is updated in an online fashion. Their work considers small updates while we explore larger changes when the target model's dataset undergoes deletion or addition of classes.
\fi
We consider the following new attack setup (for parameter $\alpha$):
\begin{description}
\item[Training] Let $f_{T_1},..., f_{T_\alpha}$ be models used in training a \emph{final} model $f_{T_\alpha}$.  The attacker receives $\ell$ samples of the form \[(x_i, f_{T_1}(x_i), f_{T_2}(x), ...., f_{T_\alpha}(x_i)).\]
  At the end of this stage the attacker outputs a model $f^{-1}_{T_\alpha}$.  It should be the case that for unseen pairs $x', y'$ the value $f^{-1}_{T_\alpha}(y')$ is similar to $x'$. 
 \item[Test/Attack] The attacker receives values $y'$ and inverts them to produce realistic biometric values $f^{-1}_{T_\alpha}(y')$.
\end{description}  Multiple works have considered attack avenues to steal models~\cite{tramer2016stealing,wang2018stealing,orekondy2019knockoff}. We review three settings when multiple models are available in Section~\ref{sec:multiple models}. We ask whether an attacker who sees the output of multiple models when training the attack model is able to invert more effectively.  
The research question of this work is:
\begin{quote} How to effectively use multiple models to reduce the training set size?\end{quote}

\noindent
We consider training set size of $\ell =2\times 10^3$.  Mai et al.~\cite{mai2018reconstruction} used $\ell = 2\times 10^6$, Ahmad and Fuller~\cite{ahmad2020resist} used $\ell = 2\times 10^4$.

Our attacks are performed on raw templates which are output from biometric networks and stored insecurely. There are two relevant lines of work on securing biometric models. One line shows how to encrypt the output of biometric networks~\cite{ratha2001enhancing,zuo2008cancelable,gomez2016unlinkable,bringer2015security,stokkenes2016multi,dodis2008fuzzy,juels1999fuzzy,juels2006fuzzy,jin2017ranking,boyen2004reusable,fuller2013computational,hernandez2009biometric,bringer2007optimal,keller2020fuzzy,canetti2021reusable} in a way that authentication systems still work. These methods have constraints where the provided security (in bits) is small or authentication is slow.
 A second line show how to securely train models and allow these models to be evaluated privately~\cite{mohassel2017secureml,yang2019federated,rouhani2018deepsecure}.  Our attacks are black box but do need the ability to observe $f_{T_i}$ for multiple $i$. 

\subsection{Attack Approach}
The high level architecture of our inversion attack is a generative adversarial network or GAN~\cite{goodfellow2014generative} as in prior work on biometric model inversion~\cite{mai2018reconstruction,ahmad2020resist}.  A GAN is a pair of algorithms, a generator and a discriminator.  In usual image applications, the generator takes random noise.  The generator's goal is to produce images that the discriminator cannot distinguish from true training samples.  As with previous work~\cite{mai2018reconstruction,ahmad2020resist}, we modify this paradigm, making the GAN generator take the output of biometric transform as input.  The discriminator is then given either real biometrics or those created by the generator.  By fooling the discriminator, the generator works as our attack model and an inverter for the biometric transform. 
Yang et al.~\cite{yang2019neural} proposed a simple mechanism for incorporating multiple models:




\begin{description}[labelindent=0cm]
\item[Random] During attack model training, a random $1 \le i \le \alpha$ is selected and the pair $(x_j, f_{T_i}(x_j))$ is provided as ground truth for the GAN.
\end{description}

\begin{figure*}[t]

	\centering
	\includegraphics[scale = 1]{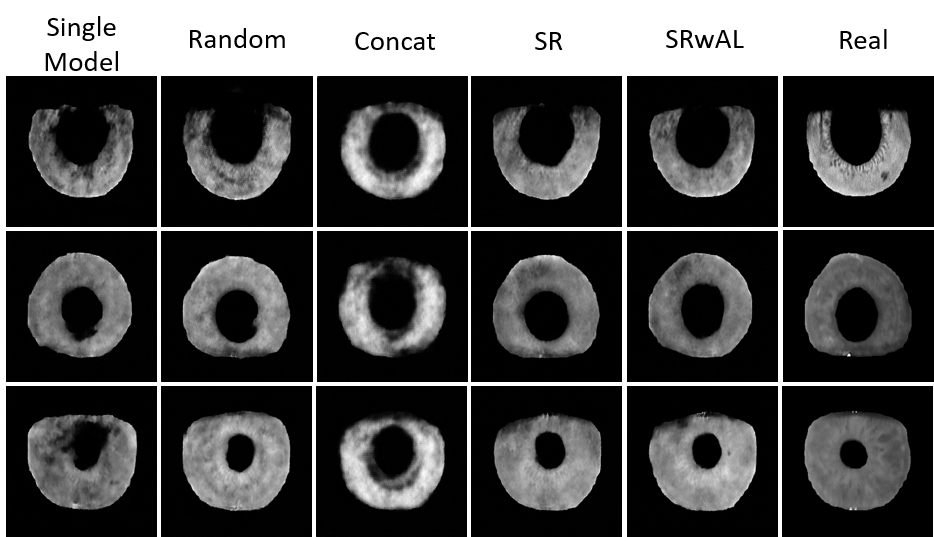}
	\caption{Visual improvement for our proposed alignment method on feature extractors. Rows represent different irises.  Columns indicate the method used as described in Introduction and Section~\ref{sec:incorporating multiple}.  All multiple model results use \upslope attack described in Section~\ref{sec:multiple models}.}
	\label{fig:alignment_improvement}
	
\end{figure*}

We show visual reconstructions of irises in Figure~\ref{fig:alignment_improvement}, deferring discussion of results and visual reconstructions of faces until Section~\ref{sec:evaluation}. The \textbf{Random} or \rand method does recover the high level shape of the iris but is missing crucial details such as 1) a crisp boundary between the iris and the pupil and 2) iris texture. 

\subsection{Our Contribution}
\label{sec:contribution} 
Let $m$ denote the output dimension of $f_{T_i}$.  Yang et al.~\cite{yang2019neural} consider a GAN with input length of $m$.  
All of our new methods consider a GAN takes inputs of length $\alpha \cdot m$.  We call these networks \emph{input-augmented GANs}.  We introduce the following input-augmented GANs:

\begin{description}[labelindent=0cm]
\item[Concatenation] In this approach the GANs training samples are the entire tuple $(x_j, f_{T_1}(x_j), f_{T_2}(x), ...., f_{T_\alpha}(x_j)).$
\item[Structured Random] Sample a random $1\le i \le \alpha$ as above and set other components of the vector to $0$.  That is, input \[(x_j, 0, 0, ..., 0, f_{T_i}(x_j), 0, ..., 0).\] 
\item[Structured Random w/ Alignment Loss] This approach follows the structured random approach above, but also asks the GAN to predict $i$.  
\end{description}

\noindent
Going forward we refer to these three methods of incorporation as $\concat, \srand$ and $\sral$ respectively. We consider two types of target models: feature extractors and classifiers (see Section~\ref{ssec:metrics}).  
 Figure~\ref{fig:alignment_improvement} shows examples of irises reconstruction from the output of a feature extractor  using  the different incorporation methods. 

\sral provides the best results see visual reconstructions in Figure~\ref{fig:alignment_improvement} and detailed results in Section~\ref{sec:results}).  This is interesting in comparison to \srand because the only difference is that  \sral asks the model to remember which location $i$ is nonzero. Even though the value $i$ is ``easy'' to predict, forcing the GAN to predict this value improves overall performance.  We believe that the GAN is better able to distinguish between inputs from different models, which leads to better inversion on the final model $f_{T_\alpha}$.
Our accuracy results are in Tables~\ref{tab:reconstruct slopes} and ~\ref{tab:reconstruct}.


\ifnum\membershipinference=1
\subsection{Application to Membership Inference} We consider a secondary application to \emph{membership inference}~\cite{shokri2017membership}.
\ifnum\conference=1
\footnote{There are many different attacker postures for such attacks (see Papernot et al.~\cite{papernot2018sok} for an overview), in this work we make the standard assumption that the adversary has query access to the target model and part of the dataset that was used in training the target model.}
\else
There are many different attacker postures for such attacks (see Papernot et al.~\cite{papernot2018sok} for an overview), in this work we make the standard assumption that the adversary has query access to the target model and part of the dataset that was used in training the target model.
\fi 
  The goal of membership inference is to determine whether a particular sample $x$ was used in the training of a network.  That is, given 
$
f_{T_\alpha}(x_i)
$ decide if $x_i$ was used to train the model $f_T$. 

Inferring whether a sample was used in training a network has privacy implications, allowing an attacker to infer a user's race, gender, or their ability to access a system.  

Membership inference attacks are usually conducted on classification networks. Mitigation strategies include regularization methods such as dropout and weight regularization~\cite{shokri2017membership}. Nasr et al.~\cite{nasr2019comprehensive} study membership inference attacks in detail under many settings, they also consider model updates. Leino et al.~\cite{leino2020stolen} look at white box membership inference attacks and conclude that a small generalization error does not guarantee safety against attacks. Chen et al.~\cite{chen2020gan} study membership inference attacks on GANs. Melis et al.~\cite{melis2019exploiting} explore membership inference attacks with model updates.

Our attack model for inferring membership is a simple neural network with 3 layers with the last layer predicting membership of the input vector. Even with this simple network for iris feature extractors membership inference is $99\%$ accurate (Tables \ref{tab:mi slopes} and\ref{tab:membership inference_class}). Membership inference is substantially harder on classification networks than feature extractors.  

As before we compare the $\rand, \concat, \srand$ and $\sral$ methods.  Membership inference on classification networks using a single model is $52\%$ accurate, but using $\sral$ raises accuracy up to $62\%$.
\fi

\paragraph{Organization} The rest of this work is organized as follows: Section~\ref{sec:architecture} describes the system architecture, Section~\ref{sec:intro biometrics} reviews how feature extractors and classifiers are used in biometrics, Section~\ref{sec:design} describes our attack model, Sections \ref{sec:evaluation},
\ref{sec:results} present  evaluation methodology and results respectively.  
\ifnum\membershipinference=1
Section~\ref{sec:membership inference} presents both methods and results for membership inference. 
\fi
Section~\ref{sec:conclusion} concludes.



\section{Adversarial Model}
This section describes the adversarial model.
\ifnum\conference=0
 and goals of model inversion and membership inference
 \fi
We defer discussion of measuring attacker success until Section~\ref{ssec:metrics}. 
 Recall that we use $x$ to denote the input to the target network and $y = f_T(x)$ to indicate the resulting output. The goal of the attack is to train a network $f_{T}^{-1}$ that on input $y$ that can predict $x$.  As mentioned in the Introduction, we assume that the adversary has access to the output of multiple related models.  That is, in the \textbf{Training} stage they receive tuples of the form
 \[x_i, f_{T_1}(x_i), f_{T_2}(x), ...., f_{T_\alpha}(x_i).\]
 the goal of the training stage is to produce a model $f_{T_\alpha}^{-1}$ where it is true that 
 \[
 f_{T_\alpha}^{-1}(f_{T_\alpha}(x')) \approx x'.
 \]

We use $i$ to additionally index the target model, for example: 
$
     x, \{f_{T_i}(x) = y_i\}_{i=1}^\alpha. 
$
The parameter $\alpha$ controls how many models the adversary has access to.
The second stage of the attack is denoted as $\textbf{Test}$ where we assume outputs $f_{T_\alpha}(x')$ leak and the attacker will reconstruct $x'$. 

\label{sec:architecture}


\subsection{Accessing Multiple Models}
\label{sec:multiple models}
We consider three types of related models that may be available to an attacker that we call \upslope, \update and \downslope. 


\paragraph{\upslope} In the first setting, we consider the intermediate models that are created when a model is first trained.  Due to the complexity of modern models, training is a computationally intensive process and is done in epochs. Since training is a complex, error-prone process models and performance data are stored for debugging purposes.
\ifnum\conference=1
\footnote{Salem et al.~\cite{salem2020updates} considered the related question is whether the difference in two models when an individual item is added leaks about that individual item.}    
\else
Salem et al.~\cite{salem2020updates} considered the related question is whether the difference in two models when an individual item is added leaks about that individual item.
\fi
\ifnum\conference=0
Since model training is intensive and is often hand-tuned, these models and their results on training data are stored for debugging purposes.  We assume the adversary has access to these models. We assume that multiple models are used at training time.  
\fi
The target iris and face recognition models converge in 100 epochs (see Section~\ref{sec:target models}). 

We utilize five different models saved during training in this attack. We utilize models after 0 (Pre-trained on ImageNet), 25\%, 50\%, 75\% and 100\% of training.  At attack time, we consider two settings when only the \emph{final}  model's output is available and when \emph{all} models are available. The setting when all models are available at test time is used to compare the different methods for incorporating multiple vectors and is not intended to be realistic.

\paragraph{\update} 
\ifnum\conference=0
For the update attack, we assume the attacker has the ability to insert a new class that will be incorporated into training of either a classification or feature extractor network. Biometric identification systems may need to be retrained when a new user is added. 
\fi

Addition of a new user into the system that needs to be learned by the model. In this case the attacker may be able to prepare the images used in training the model on the new user. That is, the image need not come from the \emph{honest} biometric distribution.
A natural setting in which an attacker could perform \upslope and \update attacks is federated learning~\cite{yang2019federated,bonawitz2019towards}. 
 In this setting, a model is trained but the adversary asks the model to learn on new images.
 
 Images in the \update attack are crafted by the adversary. The updates are crafted by  taking a normal biometric image  and applying a Gaussian blur using a 3x3 kernel with $\sigma=0.8$ to all images of the new user being added. Blurring makes images from different classes appear similar. Fredrikson et al.~\cite{fredrikson2014privacy} perform a similar attack where they recover original faces from blurred out images however we perform a smaller amount of blur. 
  Since models are only being fine-tuned for an update we consider the following training regime for the target model.  We retrain the  target model for 10 epochs.  The attacker has access to the original, 5th and 10th model.

\paragraph{\downslope} Removal of a person from the system.  Such a removal may occur due to right to be forgotten legislation which has resulting in the field of \emph{machine unlearning}~\cite{cao2015towards,bourtoule2019machine,ginart2019making}. In this setting, the adversary requests an individual be removed but has no control over how this removal is processed.
Recent laws and regulations have also taken privacy risks into account. The General Data Protection Regulation~(GDPR)~\cite{mantelero2013eu} in the European Union and the California Consumer Privacy Act~(CCPA) \cite{palmieri2020should} in the United States call for more action to protect personal data and control how and where data is stored. In addition to simplifying rules on data storage and privacy this legislation grants control to a person over their personal data, consequently, a person can ask a company to remove their data. 
 
\ifnum\conference=0
Recent legislation (including GDPR) has given individuals control over their data and the ability to request ``to be forgotten.'' This right to be forgotten clause is hard to implement in terms of machine learning models.
One naive solution is to retrain the model on the dataset after removing required data. Some machine learning models take months of computation time to train while the dataset in use may also be large in size. This problem has realized a subfield of machine learning called \textit{Machine unlearning}~\cite{cao2015towards,bourtoule2019machine,ginart2019making}. Some works also perform unlearning by updating model parameters and not retraining~\cite{ginart2019making}. 

Taking an example of a biometric system where a model is trained on a set number of people. When an individual requests their data be deleted from the training dataset and not be part of the model the biometric system must perform machine unlearning to remove the individuals data. 
\fi 
We assume machine unlearning is performed naively: completely retraining the model after deleting required data from the training dataset.  Attacking more sophisticated unlearning strategies is an important piece of future work.

We retrain the target model~(to perform unlearning) for 100 epochs on the new training dataset. We utilize models after 0, 25\%, 50\%, 75\% and 100\% of re-training has been completed for a total of 5 models. We assume the adversary removes multiple people/classes from the training set of a model. We remove 10 classes from our iris application and 5 classes from our face dataset and then retrain the model.

For all attacks except for the \update attack, the adversary can passively  receive normal biometric images and their corresponding outputs. As mentioned above for the \update attack, these images are prepared specially and differ from the normal biometric distribution.

In our attacks we only assume the output of the model, either a template or a classification vector, is revealed. This is in contrast to models that assume knowledge of the internal weights of the models $f_{T_i}$ such as Fredriskson et al.~\cite{fredrikson2014privacy}.


\section{Review of Types of Biometric Target Models}
\label{sec:intro biometrics}
	\textbf{Feature extractors $f_{E, T}$} Feature extraction networks~\cite{Goodfellow-et-al-2016} output a $m$-dimensional feature vector.  Feature extractors are trained to generate embeddings from given biometric inputs. As an example, given a biometric input $x$ a network $f_{E, T}$ generates an embedding/feature vector $y = f_{E, T}(x)$. This embedding is known as a \emph{template} and stored in a database (in mobile devices this storage is inside of a secure enclave).  At a subsequent reading of the same biometric input (with noise) another embedding $y' = f_{E, T}(x')$ is produced. A distance metric such as Euclidean distance is used to compare the two vectors $d = L_2(y, y')=\sqrt{\sum_{i=1}^n (y_i -y_i')^2}$ . The biometric will authenticate an individual successfully if the distance is below a precomputed threshold, denoted $\mathtt{thres}$. Training feature extraction networks generally does not require a set number of classes, only labeling which samples should be grouped together or pushed apart. Feature extractors are used in applications when not all users are known when the model is trained.  

	\textbf{Classifiers $f_{C, T}$} Classification networks output an $m$-dimensional classification vector.  Classifiers work with known number of classes. The objective is to learn a classification vector such that every input $x$ that belongs to a class from $\{1, \dots, m\}$ is assigned to its class in the classification vector. Usually, the output layer of a classification network is a softmax based layer which takes the preceding (feature vector) layer and maps it to a classification vector. When all users are known at training time, classifier use for biometric identification is straightforward.  A biometric is deemed to belong to class $i$ if the classification output indicates membership in class $i$ with high enough confidence (which depends on the application).  

While classifiers and feature extractors have similar network architectures, feature extractors are expected to identify \emph{new} individuals that were not seen during training.

\subsection{Attack Goal}
\label{sec:model inversion}
As a reminder, for a target network $f_T$ where $y' = f_T(x')$ the goal is to learn a transform $f_T^{-1}$ such that for $x^*= f^{-1}_T(y')$ the values $f_T(x')$ and $f_T(x^*)$ are similar.

We briefly review the goals for the two settings of feature extractors and classifiers.  
 For feature networks the goal is given $y = f_T(x)$ to produce an $x'$ such that $x'\sim x$.  In classification networks, for class $i$, the goal is to produce an $x'$ that will be classified as class $i$ with the highest confidence possible~\cite{shokri2017membership}.  As we are using a GAN to produce these $x'$ there is a secondary goal that $x'$ appears similar to valid $x$. This may not be the case if $x'$ was simply the \emph{class average}~\cite{yang2019neural}.
 
 The reason for the difference in goal is because of the difference in how these network types are used in identification systems.  Feature extractors and classifiers are used differently in identification systems.  Feature extractors are used to extract templates that are compared with a stored value.  Thus, the goal is to be able to recreate the stored template as accurately as possible.  Classifiers judge an input to be in a class if has ``high enough'' confidence of being assigned to that class so the goal is simply to maximize that confidence. The attacker goal in both settings to produce an image that will \emph{authenticate} with the highest probability. In the literature the feature extractor inversion task is called \emph{reconstruction}~\cite{ahmad2019thirdeye} while the classifier inversion task is called \emph{model inversion}~\cite{mai2018reconstruction}.  We do both in this work. To summarize the goals of model inversion are as follows:
 \begin{description}
 \item[Feature Extractor] Given $y = f_T(x)$ find $x'$ that is similar to original $x$,
 \item[Classifier] For class $i$, find $x'$ that is labeled $i$ with high confidence and cannot be distinguished from a real image.
 \end{description}

\noindent
In membership inference, the goal is the same in both cases, given $y$ determine if $y$ was part of the training set.

\section{Attack Model Design}
\label{sec:design}

\ifnum\conference=0
Since we utilize models at different stages of training, their accuracy on the training
and testing dataset will be different. This difference will show in the vectors obtained by querying the target models. The goal is to boost
\ifnum\conference=0 membership inference and
\fi model inversion
accuracy by utilizing these additional vectors per input instead of just a
single vector. 
\fi

We assume black-box access to the target network. The attack setting defined by Shokri et al.~\cite{shokri2017membership} for membership inference attacks uses shadow models trained on a shadow dataset to mimic the target model. These shadow models are then used to generate training data for attack models ultimately used to perform inference attacks.

The adversary is assumed to have  an attack dataset that comes from the same distribution as the target models training data~\cite{salem2018ml}. We utilize the same assumptions as the original membership inference attack~\cite{shokri2017membership} except we attack the target model directly (having black-box access) without generating shadow models. We also assume the system training the target model saves models at each epoch.
In all three  attack settings ($\upslope$, $\update$, and $\downslope$) new models are generated which can be coupled with old models and more accurate attacks can be conducted. Since we do not train shadow models, we relax the assumption of the adversary having a disjoint training dataset.

\ifnum\conference=0
 \begin{figure*}[t]
	\centering
	\includegraphics[scale = 0.7]{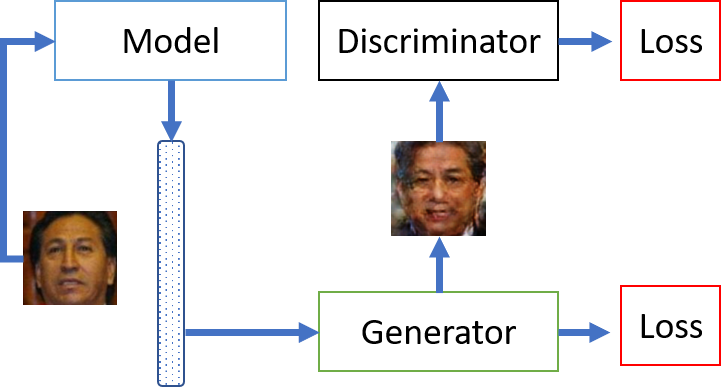}
	\caption{Inversion attack network. A biometric recognizing model is queried with biometric images to generate vectors. These vectors are fed to the generator which reconstructs an image from the vector.}
	\label{fig:gandiagram}
	\vspace{-7mm}
\end{figure*}
\fi





\subsection{GAN design}
Our attack network is a GAN~\cite{goodfellow2014generative}. A GAN architecture has two sub-models, a generator which generates images and a discriminator which judges how good the generated images are.
\ifnum\conference=0
 This is shown visually in Figure~\ref{fig:gandiagram}.  
 \fi
 Usually, the input of the generator is a noise vector sampled from a multivariate normal distribution. In our attack case the generator of the inversion attack model  is an autoencoder which takes input a feature vector (or a prediction vector) and tries to reconstruct the corresponding image by minimizing multiple loss functions. The core of prior biometrics model inversion attacks is also a GAN~\cite{mai2018reconstruction,ahmad2020resist}.

The discriminator and generator of a GAN model can be summarized in two loss equations:

\begin{align}
\ifnum\conference=1
L(D) =& -\mathbb{E}_{x \sim \mathbb{P}_{real}} [\log(D(x))]\nonumber\\&-\mathbb{E}_{x' \sim  \mathbb{P}_{{fake}}} [\log(1 - D(x'))] \\
\else
L(D) =& -\mathbb{E}_{x \sim \mathbb{P}_{real}} [\log(D(x))]-\mathbb{E}_{x' \sim  \mathbb{P}_{{fake}}} [\log(1 - D(x'))] \\
\fi
L(G) =& -\mathbb{E}_{x' \sim \mathbb{P}_{fake}} [\log(D(x'))].
\end{align}

 Where $L(D)$ is the discriminator loss and $L(G)$ is the generator loss. The variables $x$ and $x'$ correspond to the original and inverted image respectively. The discriminator's loss function is the difference in its classification performance for original and inverted images.  The generator's loss function is how well the discriminator does in identifying fake images.  
 
 For feature extractor target networks, the GAN model takes as input a feature vector. For classification networks, the GAN takes as input a classification vector. Recall that these two attacks have different goals, the feature extractor GAN is trying to reproduce $x'$ as accurately as possible.  The classifier GAN is trying to find an instance $x'$ that will be assigned to the appropriate class label with the highest confidence possible. We do not address explicitly train our models to generate $x'$ samples which will have a high inversion attack accuracy.  Instead, minimization of visual difference between original and reproduced samples is a proxy for inversion attack accuracy.

The generator loss functions include the L1 loss, SSIM~\cite{wang2004image} loss and the perceptual loss~\cite{johnson2016perceptual} between the inverted and actual image. We minimize: 1) the L1, 2) the perceptual loss, and 3) the structural dissimilarity (or maximizing structural similarity). Finally since a GAN consists of a generator and a discriminator, the generator is fine tuned by the output of the discriminator. Our final objective for the generator including the discriminator loss is:
\begin{equation}
L_{G}=  L_{Perceptual} +  L_{L1} +  L_{SSIM} + L_{D}
\end{equation}

Where $L_{D}$ is the discriminator output affecting the generator along with other reconstruction loss function as in~\cite{ahmad2020resist}.



\subsection{Incorporating multiple vectors}
\label{sec:incorporating multiple}
 There are multiple ways the additional vectors (for both attack types) per image can be used to better train our attack models. 
 
 In the \rand approach, for every update to our attack models we sample vectors randomly (with a probability of $1/ \alpha$) chosen from the outputs of one model among $\alpha$ models. The inversion model then learns to invert these feature vectors to their corresponding images. The inference attack model learns to differentiate between training and non-training samples.
 
 \paragraph{Augmented GAN mechanisms}
 Merging vectors to form a long vector is another way of feeding additional information to our attack models. 
 
 The \concat method takes $\alpha$ vectors of size $m$ to form an input vector of size $\alpha \cdot m$. The attack now learns from multiple models in one training step.
The structured random or \srand approach, we randomly sample a vector as in our random approach but instead of a $m$ sized vector we form a $\alpha \cdot m$ sized vector with all zeros except the randomly sampled vector placed in the $i^{th}$ index : \[0, 0, ..., 0, f_{T_i}(x_j), 0, ..., 0.\] Intuitively, we force the inversion model to differentiate between vectors gathered from multiple models. This enables the inversion model to learn how the output of a target model changed as it trained (or untrained) to convergence. The attack model now learns from a single vector in a single learning step while having the context of multiple vectors across multiple learning steps.

The structured random w/ alignment loss or \sral forces the attack model to predict the $i^{th}$ index or the index which holds the non-zero vector. 
In the \sral method we add to the GAN an additional loss 
    $L_{A} = \sigma({z})$ where $\sigma$ represents the softmax function and cross-entropy loss applied on an intermediary layer $z$ in the generator model. This layer predicts the index of the randomly chosen vector. This prediction forces the generator model to implicitly learn features from multiple vectors extracted from multiple models.
    
\begin{figure*}[t]
	\centering
	\includegraphics[scale = 0.60]{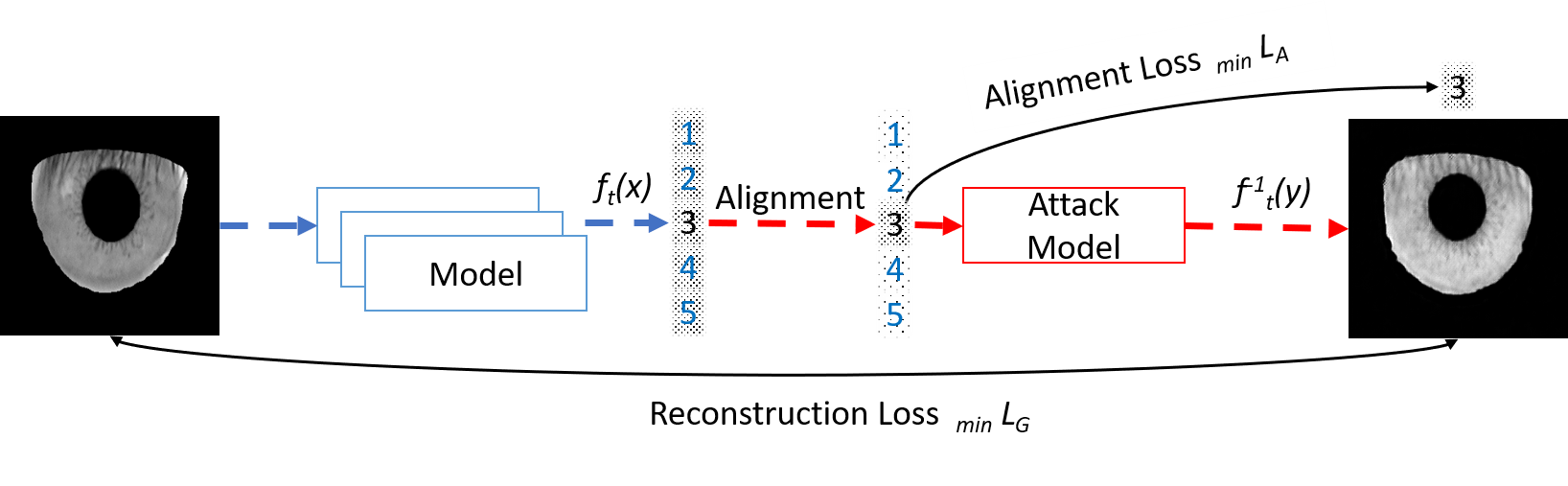}
	\vspace{-.2in}
	\caption{Vector alignment process for \sral method of incorporating multiple vectors. When reconstructing from vectors only a single feature vector is used while the rest are truncated to zero. The inversion model now implicitly learns from multiple vectors over the entire training process. }

	\label{fig:alignment}
\end{figure*}

This forces the model to further differentiate between vectors from multiple models by forcing the attack model to pass index information across its weights. Alignment loss allows the attack model to better understand how a target model was trained.  We show the setup for \sral in Figure~\ref{fig:alignment}.

\subsection{Measuring Success}
\label{ssec:metrics}
We use two standard accuracy metrics that will be used in this work for feature extractors~\cite{mai2018reconstruction,ahmad2020resist}.   
\begin{description}[labelindent=0cm]
\item[Rank-1] How frequently the inverted biometric value $x^*$ is closest to a biometric from the same class \textbf{excluding the reading used to invert}. A true positive for rank-1 accuracy is when the reconstructed image's extracted feature vector is closest to a feature vector belonging to a member of the same class as the target image. For a set of different biometrics $\mathtt{Bio} = \{\mathtt{Bio}_j\}$ consisting of pairs $x_{i,j}, y_{i,j} = f_T(x_{i,j})\in \mathtt{Bio}_j$ a true positive is when
\[
\argmin_{(x^*, y^*)\in \mathtt{Bio}, y^* \neq y_{i,j}}\left\{d(y^*, f_T(f^{-1}_T(y_{i,j})) )\right\} \in \mathtt{Bio}_j
\]

\item[Type1] Type1 considers the quality of biometric with respect to a specific distance threshold $t$.  That is, we first compute $t$ as the maximum value such that the false accept rate (FAR) of an image of a different biometric (in the underlying $f_T$) is at most $.01$ on the target model's training dataset.  It then considers how frequently the reconstructed image produces a feature vector that would be \emph{accepted} by a system with threshold $t$.    Mathematically, this is written as 
\[
\Pr
\ifnum\conference=0
_{\substack{\mathtt{Bio}_j\in \mathtt{Bio},\\x_{i,j}, y_{i,j}\in \mathtt{Bio}_j}} 
\fi
[d\left(y_{i,j}, f_T\left(f_T^{-1}(y_{i,j})\right)\right)\le t].
\]
\end{description}

\noindent
Rank-1 accuracy is more instructive for applications with all to all matching while Type1 accuracy is more important for a spoofing application where one wishes to break into a biometric authentication system.  

For classification networks, we consider traditional accuracy:

\begin{description}
\item[Accuracy] for $y$, how frequently is $f_T(f^{-1}_T(y))$ labeled with the same class as $y$.
\end{description}

In all attacks, we do not use any images used to train the target model in the attack. Instead, we probe the target model with a \emph{probe} dataset~\cite{shokri2017membership} that is smaller than the  training dataset. This probe dataset is class disjoint from the training dataset for the feature extraction setting.


\section{Evaluation}
\label{sec:evaluation}
This section details the datasets used in training both the target models and the attack, specifies the training methodology for the target models, and describes accuracy metrics used for the attacks.

We utilize two datasets in our work, one for the iris and one for faces.
The \textbf{ND-IRIS-0405}~\cite{bowyer2016nd,phillips2008iris} dataset contains 64,980 iris grayscale images from 356
subjects. The classes are highly imbalanced, some classes have many more images than others.
Left and right irises of an individual are treated as different classes~\cite{daugman2004iris}, resulting in 712 classes.

\textbf{Labeled Faces in the Wild} (LFW)~\cite{LFWTech} face recognition dataset contains 13233 images of 5749 people downloaded from the various websites with 1680 people having 2 or more images. For our evaluation we only consider people with more than 15 images yielding 89 classes with $3482$ images.

\subsection{Target models training}
\label{sec:target models}
Our target models use the DenseNet-169 architecture from the original DenseNet paper~\cite{huang2017densely}. We use loss function from SphereFace~\cite{liu2017sphereface} coupled with the Adam optimizer~\cite{kingma2014adam} to train the target networks using Tensorflow~\cite{abadi2016tensorflow}. Dropout~\cite{srivastava2014dropout} has been studied in literature as a defense against membership inference attacks~\cite{salem2018ml}. We train our networks with dropout applied to the fully connected layer which is the second to last layer of our target classification network. The dropout ratio used is 0.5. DenseNets provide near state of art recognition accuracy when coupled with dropout. Our target networks are thus generalized and possess some defense through the use of Dropout. Mai et al.~\cite{mai2018reconstruction} and Ahmad et al.~\cite{ahmad2020resist} do not use any dropout in their target networks. 

We now discuss the training and probe dataset splits for our feature extraction and classification networks.  Our attacks against feature extraction networks the target data's training is \emph{class disjoint} from the probe attack dataset. This is not the case for classification networks which are designed for a predefined set of classes.

\paragraph{Iris - Feature Extraction}


The target model for the iris dataset is trained on left iris images of all (356) subjects forming a private training set of roughly 10000 images. We assume a probe dataset of 2000 images from right irises of 40 subjects. 

\paragraph{Iris - Classification}
We train our target model on left iris images of all subjects. The total number of images is 10000 with training done on 7000. The remaining 3000 left iris images from these subjects are used to form the probe dataset. 

\paragraph{Face - Feature Extraction}
Of the $89$ classes with $3482$ images, $39$ classes and $1500$ images are used as probe images. The target model for the face dataset is trained on the remaining $50$ classes and $1982$ images.

\paragraph{Face - Classification}
The target model is trained on all entire 89 classes leaving out 15\% images from each class to make the probe dataset.

The iris images are segmented~\cite{ahmad2018unconstrained} to not include any additional texture besides that of the iris. The reconstruction attack model therefore is forced to learn texture information stored in the output feature vector. We utilize deep-funneled images~\cite{Huang2012a} for LFW dataset and crop the images to a size of 128x128 to include the face area only.


\section{Results}
\label{sec:results}
\ifnum\membershipinference=1
We perform two attacks in multiple configurations on two biometric datasets to test the efficacy of our proposed pipelines.
\fi

\subsection{Feature Extraction Networks}
\begin{figure}[t]
	\centering
	\includegraphics[scale = 0.90]{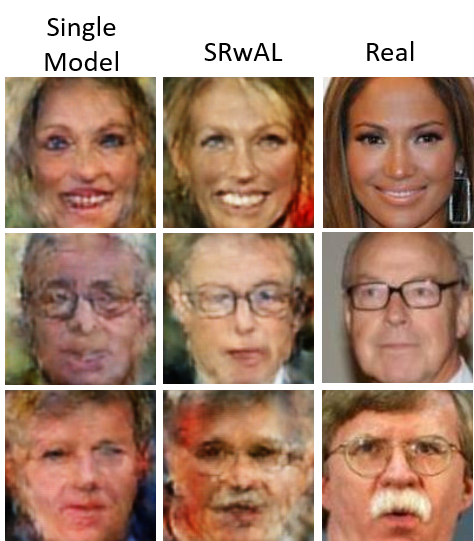}
	\caption{Alignment helps with reconstructing face features such as facial hair and correcting skin tone.}
	\label{fig:face_multiple}
\end{figure}

We show Type-1 and Rank-1 accuracy for our attacks. An overview of results in Tables~\ref{tab:reconstruct slopes} and~\ref{tab:reconstruct}.  Figure~\ref{fig:alignment_improvement} showed visual results for the iris.  Visual results for the facial biometric are in Figure~\ref{fig:face_multiple}.

\paragraph{Single Model Results}
Type-1 attack accuracy when inverting feature vectors using access to a single target model is 59\% and 85\% for the iris and face dataset respectively. 
In the Type-1 setting a reconstructed biometric is matched with its original counterpart, we obtain Type-1 attack accuracy numbers of $59\%$ compared to Ahmad and Fuller~\cite{ahmad2020resist} who achieve $75\%$ while using $10$ times the training set size.  These results are shown in Table~\ref{tab:reconstruct slopes}.

\begin{table*}[t]
	\centering
	\begin{tabular}{| l | l | r | r|   r | r | r|}
		\hline
                &         Types of&           &   Training    & \multicolumn{2}{c|}{$f_E$} & $f_C$\\
		Dataset &  Models  & \# &  set size & Type1&Rank-1 & Acc.\\
		\hline
		
		\multirow{5}{*}{ND} & Single &1 &2000& 59\%&35\% & 81\% \\
		
		& \upslope &5 & 2000  & 65\%&45\% & 82\%    \\
		
		& \update	&3 &2000    &61\%&38\% & 81\%	 \\
		
		& \downslope & 5&2000  &60\%&44\% & 82\%	 \\

		& \cite{ahmad2020resist}& 1&20000 &75\%&96\% & -	\\
		\hline
		
		\multirow{5}{*}{LFW}& Single&1 &1500&85\%&82\% & 74\%   \\

		& \upslope&5 & 1500  & 89\%&84\%  & 78\%  \\

		& \update 	&3 &1500  & 87\%&84\%	& 75\% \\

		& \downslope & 5&1500 & 87\%&83\%	& 73\% \\

		& \cite{mai2018reconstruction} & 1&$2000000$ & 99\%&-	&- \\

		\hline
	\end{tabular} 
	\caption{Comparison of Accuracy when using multiple models with the \rand method of incorporation. Model Inversion for both Feature Extraction Networks, $f_E$ and Classification Networks $f_C$. Accuracy per dataset and attack type.  Accuracy for classification networks is how frequently an image is assigned the correct class label.}
	\label{tab:reconstruct slopes}

\end{table*}

\begin{table*}[t]
	\centering
	\begin{tabular}{| l | l | r | r|l|r|r|r|}
		\hline
                &         Incorporation&            &  Training           & Models& \multicolumn{2}{c|}{$f_E$} & $f_C$\\
		Dataset & Method  & \# Models& set size & for Test & Type1&Rank-1 & Accuracy\\
		\hline
		\multirow{6}{*}{ND} 		 & \rand &5 & 2000 &Final & 65\%&45\% & 82\%    \\
		& \concat  & 5&2000  & Final &48\%&27\% & 78\%	 \\
		& \concat  & 5&2000  & All &50\%&30\% & 86\%	 \\
		& \srand & 5&2000 & Final & 66\%&46\% & 81\%	 \\
		& \sral & 5&2000 & Final & 72\%&53\% & 83\%	 \\
		& \sral & 5&2000 & All& 65\%&45\% & 81\%	 \\
		& \cite{ahmad2020resist}& 1&20000 &$-$&75\%&96\% & -	\\
		\hline
		\multirow{6}{*}{LFW} &\rand&5 & 1500 & Final & 89\%&84\%  & 78\%  \\
& \concat & 5&1500 & Final & 78\%&75\% & 74\%	 \\
		& \concat & 5&1500 & All & 80\%&78\% & 81\%	 \\
& \srand & 5&1500 & Final & 89\%&84\% & 79\%	 \\
& \sral & 5&1500 & Final & 91\%&86\% & 79\%	 \\
		& \sral & 5&1500 & All & 89\%&84\% & 79\%	 \\
		& \cite{mai2018reconstruction} & 1&$ 2000000$ & $-$& 99\%&-	&- \\
		\hline
	\end{tabular} 
	\caption{Comparison of Methods for incorporating multiple models.  All data uses \upslope models. Both Feature Extraction Networks, $f_E$ and Classification Networks $f_C$. Accuracy per dataset and attack type.  Accuracy for classification networks is how frequently an image is assigned the correct class label.  Models for Test Column indicates whether all models or just the final model were used during testing.}
	\label{tab:reconstruct}
\end{table*}

Rank-1 accuracy measures the probability of a reconstructed biometric being matched with an original biometric of the same class~(and not itself).
Our Rank-1 accuracy is lower. Our inversion network seems to do better at the specific task of inverting a template to a particular image and does not generalize well. We attribute this to 1) slight overfitting of the inversion network due to our small training dataset and 2) 
 differences to the underlying target network in comparison to the target network of Ahmad and Fuller~\cite{ahmad2020resist}.  These differences include a more modern loss function and the use of dropout.  Additionally, our target network Rank-1 accuracy on the test set of $98.2\%$.  Ahmad and Fuller used a target network with an accuracy of 99.5\%. This accuracy changes the threshold distance used to accept or reject biometric comparisons; this change affects Type1 attack accuracy but not Rank-1.  This larger distance threshold may explain the relatively high Type1 accuracy.

This split between Type1 and Rank-1 is not observed for the facial task. We achieve a Type-1 accuracy of 85\% when using a single model to obtain the training dataset for the inversion attack network. Our inversion attack network achieves ~85\% accuracy in Type-1 and ~82\% in  Rank-1 settings. Face images have myriad of facial features in addition to some background of the LFW images making them easier to invert and harder for the target model to achieve high test accuracy. Iris images have only the iris texture while other features such as the skin and eyelid are segmented out.

\paragraph{Incorporating Multiple Models}
Turning to the setting of multiple models, we present the gain in using multiple models with the \rand technique in Table~\ref{tab:reconstruct slopes}.  In all settings, multiple models improve accuracy of the inversion network.  Because all attack settings perform similarly, in comparing how to effectively incorporate multiple models we focus on the \upslope model.

Results for different incorporation techniques are presented in Table~\ref{tab:reconstruct}.  The largest gain is using the technique \sral. For the iris, this technique boosts Rank-1 accuracy from $45\%$ to $53\%$. \textit{Input-augmented GANs} boost attack accuracy in most settings. 

The \srand technique performs nearly identically to the \rand technique.  This is of particular interested compared to the \sral technique which is only forcing the model to learn the provided input which is \emph{easy} to predict.

\textbf{Discussion}
The \concat technique can hurt performance when only a single model output is available at testing. The most natural explanation is that  feature vectors from models which have not converged hold information that is hard to use by our inversion models.
However, if one assumes that the adversary sees the output of all models at test time, that is the adversary sees $f_{T_1}(x),..., f_{T_\alpha}(x)$ at test time this accuracy improves.  This indicates that the problem may be the mismatch between the format of the training and testing data. We note that this phenomenon is switched on \sral, providing all vectors at test time is harmful.  This supports the hypothesis that Structured Random with Alignment loss is superior for natural attack scenarios.

The accuracy gain when using multiple models is less pronounced for the LFW face dataset. In this setting, we believe that the small amount of training data resulting in the attack model overfitting the training data. However, our attack achieves close to state of art inversion accuracy while using orders of magnitude less training data.

\subsection{Classification Networks}
Model inversion attacks on classification networks output the \emph{average} of a certain class (see discussion in Section~\ref{sec:model inversion}). The attack is successful if the reconstructed biometric images are classified to their correct class by the target model.  


\paragraph{Single Model Results} Our inversion attack models perform at 81\% and 74\% attack accuracy for the iris and face datasets respectively. Results are displayed in Table~\ref{tab:reconstruct slopes}. 

\paragraph{Incorporating Multiple Models} Structured random with alignment loss bumps the accuracy to 83\% and 79\% respectively.  We do not see a proportional increase in inversion accuracy as we saw with feature extraction networks. Classification networks output prediction vectors which are simple and do not hold much information. Previous works have even truncated prediction vectors~\cite{yang2019neural} for better inversion.

If all models' output is available at test time \concat method improves but the \sral does not. This same phenomenon was observed in feature extraction network. 

\begin{figure}[t]
	\centering
	\includegraphics[scale = 0.65]{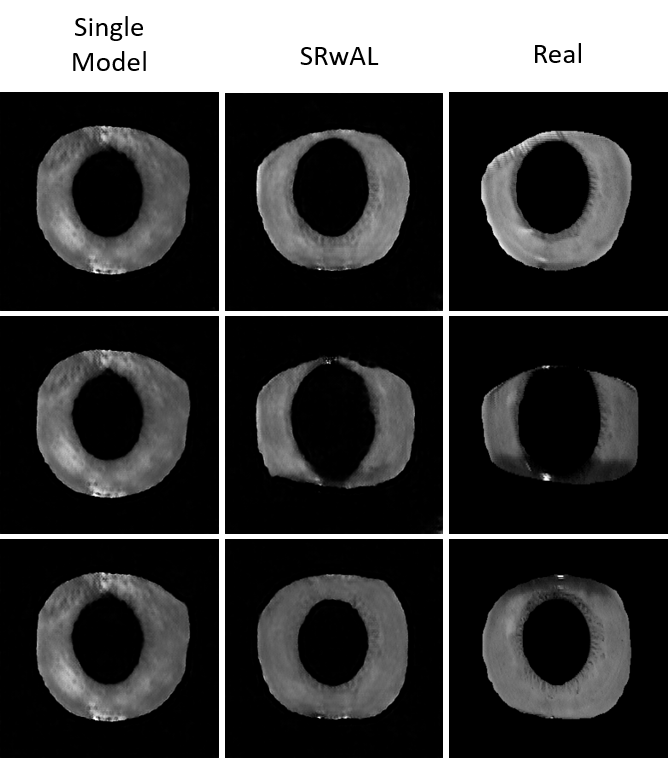}
	\caption{Alignment process enables inversion to vary. Each row represents a different iris from the same biometrics. For classification networks single model always inverts to \emph{class average}.  However, \sral can invert to distinct images that better match the stored template.}
	\label{fig:pred_multiple}
\end{figure}

Recall, for classification networks the traditional goal is to output the \emph{class average}, a value $x'$ that will be assigned to class $i$ with as high probability as possible.  Prior works have not considered that this average may not appear similar to a real biometric (such as Fredrikson et al.~\cite{fredrikson2015model}). 
When training and testing with concatenated prediction vectors ($\alpha \cdot m$) our inverted images vary across a class instead of being the same class average image. An example of different images for the same iris biometric is shown in Figure~\ref{fig:pred_multiple}.  We attribute this to the additional information in multiple vectors which form the concatenated vector. A similar phenomenon is seen in the work of Yang et al.~\cite{yang2019neural} where classes unknown to the target model are inverted by a method called alignment (that differs from \sral).

\subsection{Which attack types perform the best? }
\label{ssec:helpful models}
We perform a simple experiment to validate which models contribute the most to inversion attack accuracy. With access to the final trained target model and using random sampling for training, an adversary's reconstructed iris images have a Rank-1 attack accuracy of 35\% when provided access to the $1$st, $25$th, $50$th, $75$th, and $100$th models.  
This accuracy drops to 15\% if the adversary only has access to the 25th and 50th model. Attack accuracy jumps to 36\% if the first and 100th model are used by the adversary. 

The target model is trained using an off the shelf network architecture which was pre-trained on the Imagenet dataset~\cite{deng2009imagenet}, which would generate somewhat accurate feature vectors~\cite{boyd2019deep}. Of course, the last model generates accurate feature vectors which would allow the adversary to generate good reconstructions. Since inversion accuracy is low with intermediary models yet to converge, our proposed alignment loss forces the inversion model to differentiate between vectors from multiple models.

%
%
%
%
%
%

\ifnum\membershipinference=1

\section{Membership Inference}
\label{sec:membership inference}

\subsection{Membership inference attack}
In membership inference the goal is the same for feature extractors and classifiers: given $y$ determine if $y$ was part of the training set.  This is a binary classification task.  More formally, given a target model denoted by $f_T$ and an input $x$, we query the target model with $x$ to obtain an output $y$. This will be a feature vector or a classification vector. We assume the attacker has a small dataset from the same distribution as training data and this dataset has members from the target model's training and testing dataset. 

Our attack model is a simple $3$ layer fully connected neural network with sizes 64,64,1. The last layer infers training set membership. In the special case of $\sral$ there are two output layers predicting membership and index of the input vector.
 Our target models are trained as described in Section~\ref{sec:evaluation}.

 \subsection{Results}

\begin{table}[t]
	\centering
	\begin{tabular}{| l | l | r | r|   r | r | }
		\hline
                &         Types of&           &   Training    &\multicolumn{2}{c|}{Accuracy} \\
		Dataset &  Models  & \# &  set size & $f_E$ & $f_C$\\
		\hline
		
		\multirow{4}{*}{ND} & Single &1 &2000 & 99\% & 52\%\\
		
		& \upslope &5 & 2000  & 99\% & 60\%\\
		
		& \update	&3 &2000    & 99\% & 59\%\\
		
		& \downslope & 5&2000  & 99\% & 59\%\\
		\hline
		
		\multirow{4}{*}{LFW}& Single&1 &2000& 80\% & 68\%\\

		& \upslope&5 & 2000& 76\% & 78\%\\

		& \update 	&3 &2000  & 78\% & 75\% \\

		& \downslope & 5&2000 & 77\% & 76\%\\
		\hline
	\end{tabular} 
	\caption{Membership inference attack accuracy when using multiple models with the \rand method of incorporation. Accuracy is how frequently the attack predicts correctly whether the item was part of the training set.}
	\label{tab:mi slopes}

\end{table}
Our membership inference attack accuracy numbers shown in Tables~\ref{tab:mi slopes} and \ref{tab:membership inference_class}.  Feature extractor networks for the iris are easy to classify in all settings with accuracy of $99\%$.  More classes increases attack accuracy~\cite{shokri2017membership} for prediction vectors.  For the iris classification network, Table~\ref{tab:mi slopes} shows that accuracy using a single model is $52\%$ which is improved to $60\%$ by using multiple models and further to $62\%$ by using $\sral$.  Running this attack on feature vectors should be easier since they contain more information than prediction vectors. 

\ifnum\membershipinference=1
\begin{table*}[t]
	\centering
		\begin{tabular}{| l | l | r | r|l|r|r|}
		\hline
                &         Incorporation&  \#          &  Training           & Models& \multicolumn{2}{c|}{Accuracy} \\
		Dataset & Method  & Models& set size & for Test & $f_E$ & $f_C$\\
		\hline
				
		\multirow{6}{*}{ND} 		 & \rand &5 & 2000 &Final &99\% & 60\%\\

		& \concat  & 5&2000  & Final & 99\% & 60\%\\
		& \concat  & 5&2000  & All &99\% & 66\% \\

		& \srand & 5&2000 & Final & 99\% & 60\%\\

		& \sral & 5&2000 & Final & 99\% & 62\%\\
		& \sral & 5&2000 & All& 99\% & 70\%\\\hline

		\multirow{6}{*}{LFW} &\rand&5 & 2000 & Final & 76\% & 78\%\\
& \concat & 5&2000 & Final &  82\% & 74\%\\
		& \concat & 5&2000 & All &  82\% & 84\%\\
		
& \srand & 5&2000 & Final & 82\% & 75\%\\

& \sral & 5&2000& Final & 81\% & 76\%\\

		& \sral & 5&2000 & All & 82\% & 86\%\\
		\hline
	\end{tabular} 
	\caption{Membership inference attack accuracy on both classification networks $f_C$ and feature vector networks $f_E$ for different incorporation methods. All results consider the \upslope attack setting.}
	\label{tab:membership inference_class}
\end{table*}
\fi

For the face, feature extraction networks have lower accuracy which is actually hurt by using $\rand$ to incorporate multiple models, however $\sral$ does slighly outperform a single model.  As with the iris, using multiple models to attack classification networks has a more pronounced effect on accuracy.  For the face a single model has accuracy of $68\%$, multiple models improve to $78\%$ and $\sral$ improves further to $82\%$.   The LFW dataset is a harder dataset than the iris dataset taking a longer time to converge. Models in the beginning of training do not not output useful feature vectors for membership inference attacks.

Recall that in the model inversion task providing all models as input at test time improved the performance of a model trained with \concat but hurt performance of a model trained using \sral.  As shown in Table~\ref{tab:membership inference_class} providing all models improves the performance of both methods, with a strong affect for $f_C$ for both the iris and face. We attribute this difference to a difference in the two tasks, model inversion is trying to reconstruct a full image, while model inversion is only classifying an input.  As such it seems the attack model for membership inference is better able to use information from multiple models. 

\subsubsection{Which attack types perform the best? }
As with model inversion, we perform a simple experiment to validate which models contribute the most to inference accuracy. We train our attack model on a single training model and use it to attack the final training model produced during the $100$th epoch.  As expected, when using the model output after the 1st epoch to attack the model output after the $100$th epoch accuracy is only $52\%$, increasing to $85.5\%$ when using the model from the $25$th epoch, $91\%$ for the model from the $50$th epoch, and finally $99.5\%$ when using the training using the model of the $100$th epoch (same model for training and attack).
\fi


\section{Conclusion}
\label{sec:conclusion}
An adversary can perform model inversion attacks to gain unauthorized access to biometric authentication systems through biometric spoofing. We explore an adversary's access to deep learning models trained and stored, models generated after a model is updated, and finally models generated after an unlearning request. 

In this work we show when multiple models are accessible by an adversary model inversion attacks can be performed with fewer training samples with high attack accuracy. We explore different methods of incorporating multiple models into the attack model training process.  

An interesting finding of our work is that while incorporating multiple models using the \rand method is universally helpful (across biometrics and types of biometric transforms), results using input-augmented GANs are mixed. If only the last model is available at \textbf{Test} time the \concat technique can actually hurt performance, for the iris Type1 accuracy drops from 59\% to 48\% and is much lower than the 65\% achieved by the random method. However, our proposed method of using \sral always improves performance compared to the \rand technique improving Type1 accuracy to $72\%$ compared to the $65\%$ of \rand. 

To show the promise of our augmented GAN techniques, we apply them to a secondary application of membership inference.  As with model inversion, $\sral$ performs best when only the final model is available at attack time. 

\section*{Acknowledgements}
The authors thank the reviewers for their valuable help in improving the manuscript.  This work was supported in part by NSF Grants \# 1849904 and 2141033. 
This material is based upon work supported by the Defense Advanced Research Projects Agency (DARPA) under Air Force Contract No. FA8702-15-D-0001. Any opinions, findings, conclusions or recommendations expressed in this material are those of the author(s) and do not necessarily reflect the views of DARPA.





\bibliographystyle{IEEEtran}
\bibliography{acmart}

\end{document}